\documentclass[10pt,twocolumn,letterpaper]{article}

\usepackage{cvpr}
\usepackage{times}
\usepackage{epsfig}
\usepackage{graphicx}
\usepackage{amsmath}
\usepackage{amssymb}
\usepackage{subfigure}
\usepackage{algorithm}
\usepackage{algorithmic}

\def\G{{\cal G}}
\def\N{{\rm N}}
\def\E{{\rm E}}
\def\1{{\rm 1}}
\def\Cat{{\rm Cat}}

\usepackage[pagebackref=true,breaklinks=true,letterpaper=true,colorlinks,bookmarks=false]{hyperref}

 \cvprfinalcopy 

\def\cvprPaperID{5828} 

\ifcvprfinal\fi
\begin{document}

\title{Deep Unsupervised Clustering with Clustered Generator Model}
\newcommand*\samethanks[1][\value{footnote}]{\footnotemark[#1]}
\author{Dandan Zhu$^1$,
Tian Han$^2$\thanks{Tian Han is the corresponding author},
Linqi Zhou$^3$, Xiaokang Yang$^1$, Ying Nian Wu$^3$\\
$^1$Shanghai Jiao Tong University \, $^2$Stevens Institute of Technology \, $^3$ University of California, Los Angeles\\
{\tt \small \{ddz,xkyang\}@sjtu.edu.cn, than6@stevens.edu, linqizhou907@gmail.com, ywu@stat.ucla.edu}
}

\maketitle
\thispagestyle{plain}
\pagestyle{plain}
\begin{abstract}
  This paper addresses the problem of unsupervised clustering which remains one of the most fundamental challenges in machine learning and artificial intelligence. We propose the clustered generator model for clustering which contains both continuous and discrete latent variables. Discrete latent variables model the cluster label while the continuous ones model variations within each cluster. The learning of the model proceeds in a unified probabilistic framework and incorporates the unsupervised clustering as an inner step without the need for an extra inference model as in existing variational-based models. The latent variables learned serve as both observed data embedding or latent representation for data distribution. Our experiments show that the proposed model can achieve competitive unsupervised clustering accuracy and can learn disentangled latent representations to generate realistic samples. In addition, the model can be naturally extended to per-pixel unsupervised clustering which remains largely unexplored.  

\end{abstract}

\section{Introduction}
Clustering as one of the central themes in data understanding and analysis has been widely studied in the realm of unsupervised learning. However, unsupervised clustering remains one of the most fundamental challenges in machine learning because of high dimensionality of data and  high complexities of their hidden structures. 

Long-established approaches for unsupervised clustering including K-means~\cite{hartigan1979algorithm} and Gaussian Mixture Model (GMM)~\cite{bishop} are still the building blocks for numerous applications due to their efficiency and simplicity. However, their distance metrics are limited to data space, making them ineffective for high-dimensional data such as images. Therefore, considerable efforts have been put into obtaining a good feature embedding of data, usually of low dimensionality, for effective clustering~\cite{xie2016unsupervised}. However, the representation obtained by standalone data embedding typically cannot capture the latent structure and variation of the observed data which
may be ineffective for clustering. We believe the good representation for clustering should also be able to compactly represent the observed data distribution to encode all necessary characteristics of the observation. 

Deep generative models (a.k.a the \textit{generator models}) have shown great promise in learning latent representations for high-dimensional signals such as images and videos~\cite{radford2015unsupervised,kingma2013auto,HanLu2016}. Generator models parameterized by deep neural networks specify a non-linear mapping from latent variables to observed data. As a compact probabilistic representation of knowledge, it can embed the high-dimensional data into low-dimensional latent representation. Besides, it has been shown that the generator model is also capable of generating realistic images indicating that the learned latent representation encode all necessary and useful information of the data. Though powerful, the generator model is mainly studied with the focus on generation tasks using continuous latent variables. While it is clear that we pursue both objectives of jointly learning latent representations and clustering, developing and learning such generator model for unsupervised clustering is still in its infancy with only a few recent existing works~\cite{jiang2016variational,dilokthanakul2016deep,kingma2014semi}.  

In this paper, we develop a new model-based clustering algorithm using generator model. Specifically, we propose to use the generator model with both discrete and continuous latent variables. The discrete latent variables are used to model cluster labels while continuous ones are used to model variations within each cluster. Such model is termed the \textit{clustered generator model} to emphasize the fact that it aims to achieve unsupervised clustering. By learning the clustered generator model, we naturally incorporate the unsupervised clustering as an inference step for discrete latent variables in an inner loop, and as a result, useful latent representations (i.e., discrete and continuous latent variables) and the unsupervised clustering are seamlessly integrated into a unified probabilistic learning framework. The experiments show that by learning the clustered generator model, we could achieve competitive or even state-of-art unsupervised clustering accuracy while obtaining realistic and disentangled latent representations. 



\subsection{Related Work and Contributions}
Our work is closely related to unsupervised clustering as well as learning the generator models. 

The most fundamental methods for clustering are the K-means~\cite{hartigan1979algorithm} algorithm and Gaussian Mixture Model (GMM)\cite{bishop}. K-means assumes the data are centered around some centroids and clusters are found by minimizing $l_2$ distance to the centroid within each cluster. GMM, on the other hand, assumes that data are generated by mixture of Gaussian distribution whose parameters are learned through Expectation-Maximization (EM) algorithm. Without utilizing the proper representation, these methods are ineffective in handling high-dimensional data whose underlying structure can be highly non-linear. Spectral clustering and its variants~\cite{shi2000normalized,ng2002spectral,von2007tutorial,yang2010image} further generalize the distance function for non-linear clusters, yet in general they can be computationally intensive and still result in unsatisfactory clustering on high-dimensional data. 

Generator models have received increasing attention over the past few years as they can effectively capture data distribution through latent representations. Generative Adversarial Network (GAN)~\cite{goodfellow2014generative} and Variational Auto-encoder (VAE)~\cite{kingma2013auto,rezende2014stochastic} are two notable examples. These generative models have shown their great potential in various applications such as image generation~\cite{radford2015unsupervised,arjovsky2017wasserstein,han2019divergence}, image completion~\cite{HanLu2016,han2018divergence}, and disentangled latent representation~\cite{higgins2017beta,chen2016infogan,han2018learning}. However, integrating such powerful knowledge representation tool with the unsupervised clustering task has not been thoroughly investigated.

Only a few existing works jointly consider learning the latent representation for data and the clustering task. Conditional-VAE (CVAE)~\cite{kingma2014semi} considers discrete latent variables for clustering and is closely related to our work, but it is primarily developed for supervised/semi-supervised learning where (part of) the data label is given. HashGAN \cite{Cao_2018_CVPR} is a novel model that combines pairs of conditional Wasserstein GAN (PC-WGAN) and hash encoded information. It mainly uses a new PC-WGAN conditional on pairwise similarity information to generate an image that is closest to the real image. However, this method is also mainly used in supervised/semi-supervised tasks. Variational Deep Embedding (VaDE)~\cite{jiang2016variational} and Gaussian Mixture Variational Auto-encoder (GMVAE)~\cite{dilokthanakul2016deep} combine GMM models and VAEs for unsupervised clustering. Adversarial Auto-encoder (AAE)~\cite{makhzani2015adversarial} can also be adapted to unsupervised clustering, but it needs to use GAN to match the aggregated posterior of latent representation with the prior of VAE, requiring complex computation and additional network structures. Other related models include Deep Embedded Clustering (DEC)~\cite{xie2016unsupervised} and more recent Invariant Information Clustering (IIC)~\cite{ji2019invariant} which specifically learn feature representations for clustering tasks. The latent representations learned by DEC and IIC are unable to represent the observed data distribution, thereby failing to generalize to other tasks (e.g., generation). While most of these variational-based models could achieve relatively impressive clustering accuracy, they need to design and learn separate inference model for cluster labels. Besides, due to the discrete nature of cluster labels, variational learning cannot take advantage of \textit{reparametrization trick} and generally need further approximation. 
 
In contrast to recent models that use variational learning for latent representation and clustering, we introduce the novel clustered generator model for unsupervised clustering. Learning such model will naturally integrate the unsupervised clustering process as an inference inner loop without utilizing additional networks or any further approximation. 

Contributions of our paper are as follows: 
\begin{itemize}
\item We propose the clustered generator model for unsupervised clustering which includes discrete latent variables to model cluster labels and continuous latent variables to capture variations within each cluster. 
\item We develop a novel learning algorithm for clustered generator model in a probabilistic framework which naturally involves the unsupervised clustering as an exact inference step without any assisting models and any approximations.
\item We conduct extensive experiments to show the effectiveness of the proposed model. Specifically, our model can achieve competitive unsupervised clustering accuracy on large-scale image datasets and could get reasonably well per-pixel unsupervised clustering, a task that has remained largely unexplored before. Besides, our model can obtain disentangled latent representations as indicated by its realistic generation.    
\end{itemize}



\section{Model and Learning Algorithm}
In this section, we describe the details of the model and the corresponding inference and learning algorithm. 
\subsection{Clustered Generator Model}
Suppose $x$ be the observed data of dimension $D$. The generator model~\cite{goodfellow2014generative} assumes the observation $x$ is generated by latent variable $z$ of dimension $d$:
\begin{equation*}
z \sim \N(0, I_d); x = \G_\theta(z) + \epsilon
\end{equation*} 
where $\epsilon \sim \N(0, \sigma^2 I_D)$ is the noise and is independent of $z$, and $\G_\theta(z)$ is the top-down neural network with parameters $\theta$. In general, the latent variable $z$ is of low-dimension (i.e., $d<D$) and is learned to (1) embed the high-dimensional data $x$ in a low-dimensional latent space, and (2) represent the data distribution of $x$ through a generative model that generates realistic samples. 

Traditional generator models have been shown to be effective in image generation~\cite{radford2015unsupervised,arjovsky2017wasserstein,han2019divergence}.
However, it only deals with the latent variable $z$ that is \textit{continuous}, making it ineffective in clustering tasks which are \textit{discrete} in nature. Therefore, we propose to use the generator model with both discrete and continuous latent variables for unsupervised clustering. 

Suppose we have $K$ clusters, the observed data $x$ is now generated by not only the continuous latent variables $z$ but also the discrete latent variables $y$ of dimension $K$ which represents the cluster labels: 
\begin{eqnarray*}
z &\sim& \N(0, I_d); y \sim \Cat(\pi);\\
x &=& \G_\theta(z, y) + \epsilon
\end{eqnarray*} 
where $\Cat(\pi)$ denotes the categorical distribution with $\pi$ being the prior probability for $K$ clusters. $\epsilon \sim \N(0, \sigma^2 I_D)$ is the noise of the model and is independent of $z$ and $y$. We call such model \textit{clustered generator} to emphasize the fact that it incorporates the unsupervised clustering naturally inside its learning framework. In this way, the latent variables $z$ and $y$ are served as both observed data $x$ embedding which is for clustering and latent representation which is for representing the data distribution of $x$. A similar form has been used in~\cite{kingma2014semi}. However, the model is not developed for unsupervised clustering. Besides, the representation learned for clustering is different from the latent representation learned for data distribution which can be ineffective in both realms. We will elaborate this point in the next section and experiments. 

\subsection{Inference and Learning}
The clustered generator model defines the generation process as: $z\sim p(z), y\sim p(y), x\sim p_\theta(x|y, z)$. Therefore, the complete data model can be defined as $p_\theta(x, y, z)=p(z)p(y)p_\theta(x|y, z)$. If we observe a set of training data $\{x_i, i=1,..., n\}$ coming from the true but unknown distribution $p_{data}$, then the learning and inference of the clustered generator model can be accomplished by maximizing the observed-data log-likelihood: 
\begin{eqnarray*}
L(\theta)&=& \frac{1}{n}\sum_{i=1}^n \log p_\theta(x_i)\\
&=& \frac{1}{n}\sum_{i=1}^n \log \int_z \sum_y p_\theta(x_i, y, z)dydz
\end{eqnarray*}
The model parameters $\theta$ can be learned by gradient descent which amounts to evaluating: 
\begin{eqnarray}
\frac{\partial L(\theta)}{\partial \theta}&=&\frac{1}{n}\sum_{i=1}^n\frac{\partial \log p_\theta(x_i)}{\partial \theta}\nonumber\\
&=& \frac{1}{n}\sum_{i=1}^n\E_{p_\theta(y, z|x_i)}\left[\frac{\partial}{\partial \theta}\log p_\theta(x_i, y, z)\right]
\label{eq:likelihood}
\end{eqnarray}
However, the evaluation of expectation in Eqn.~\ref{eq:likelihood} is in general analytically intractable. For given observation example $x$, we obtain fair samples from the posterior distribution, i.e., $y, z \sim p_\theta(y,z|x)$, using \textit{Gibbs sampler} which iteratively performs the conditional sampling on latent variables $z$ and $y$, i.e., $z\sim p_\theta(z|x, y), y\sim p_\theta(y|x, z)$. 
\subsubsection{Inference on continuous $z$:}
The continuous latent variable $z$ is sampled based on posterior distribution given $y$ fixed:
\begin{equation}
z\sim p_\theta(z|x, y) \propto p_\theta(z, x, y)
\label{eq:inferz}
\end{equation}
Fair samples can be drawn using MCMC techniques like HMC or Langevin dynamics~\cite{neal2011mcmc}. Langevin dynamics is used in this work because it can help navigate the landscape of the latent space more thoroughly and effectively. Specifically, we have:
\begin{equation}
z_{\tau+1} = z_{\tau} + \delta \frac{\partial}{\partial z}\log p_\theta(z, x, y) +
\sqrt{2\delta} \N_\tau
\label{eq:langevin}
\end{equation}
where $\delta$ is the step size and $\tau$ is the time stamp for langevin inference. $\N_\tau \sim \N(0,I_d)$ is the random noise projected in each iteration. The log-joint can be evaluated as:
\begin{equation}
\log p_\theta(z,x,y)=-\frac{||z||_2^2}{2} -\frac{||x-\G_\theta(y, z)||_2^2}{2\sigma^2} + C
\end{equation}
where $C$ is the constant which does not involve $z$. Variable $\sigma$ is the pre-specified standard deviation of our model. Note that $y$ is fixed to be the currently sampled value during the learning iteration.  
It has been shown that the dynamic has the $p_\theta(z, x, y)$ as its stationary distribution. Therefore, the fair sample for $z$ from $p_\theta(z|x, y)$ can be ensured. 

In fact, from Eqn.~\ref{eq:inferz}, we can see that for given observation example $x$, the inference on $z$ amounts to finding the suitable latent representation to resemble the observation assuming it comes from a specific cluster as indicated by $y$.
\subsubsection{Inference on discrete $y$:}
The discrete latent variable $y$ is sampled based on posterior distribution given $z$ fixed:
\begin{eqnarray}
y\sim p_\theta(y|x, z)=\frac{p_\theta(x, y, z)}{\sum_y p_\theta(x, y, z)}
\label{eq:infery}
\end{eqnarray}
Suppose we have $K$ clusters, then:
\begin{eqnarray}
p(y=i) &=& \frac{p_\theta(x, y=i, z)}{\sum_{i=1}^K p_\theta(x, y=i, z)}
\label{eq:posy}
\end{eqnarray}
where
\begin{equation}
p_\theta(x, y=i, z)\propto \pi_i \exp\left[{-\frac{||x-\G_\theta(y=i, z)||_2^2}{2\sigma^2}}\right]
\end{equation}
and $\pi_i$ is the prior probability of $i$-th cluster which is pre-specified.

In fact, from Eqn.~\ref{eq:infery}, the inference on $y$ is based on true posterior distribution and essentially estimates the probability of observed $x$ falling into each cluster based on the current latent representation $z$. This is essentially unsupervised clustering based on the current representation $z$ and the model $\G_\theta(\boldsymbol{\cdot})$. Existing variational-based models~\cite{kingma2014semi,jiang2016variational} have to design and learn a separate inference model for $y$, i.e., $q_\phi(y|x)$, int order to approximate the true posterior distribution $p_\theta(y|x)$ which can be ineffective as demonstrated in our experiments.  

\subsubsection{Learning model parameter $\theta$:}
For given observed example $x_i$, after obtaining inferred continuous latent variable $z_i$ using Eqn.~\ref{eq:langevin} and discrete latent variable $y_i$ using Eqn.~\ref{eq:posy}. We then use the sampled $y_i$ and $z_i$ to learn the clustered generator model by stochastic gradient descent as in Eqn.~\ref{eq:likelihood}. More precisely, 
\begin{eqnarray}
\frac{\partial L(\theta)}{\partial \theta}
&=& \frac{1}{n}\sum_{i=1}^n\left[\frac{\partial}{\partial \theta}\log p_\theta(x_i, y_i, z_i)\right]
\label{eq:likelihood_learn}
\end{eqnarray}
The whole algorithm iterates the above three steps until convergence. Note that \cite{HanLu2016} shares the similar alternating nature as ours. However, their model does not consider the discrete latent variable and is mainly developed for image generation. See Algorithm~\ref{alg:learn} for an summarized learning and inference of our model.  

Note that the whole algorithm can be efficient and scale well for relatively large datasets which can be shown in our experiments. Though we use the Langevin sampling on $z$ which involves multiple steps, however, the gradient in Eqn.\ref{eq:langevin} shares the same chain rule computation as in Eqn.\ref{eq:likelihood_learn} which greatly reduce the computation burden. 

\begin{algorithm}[t]
\caption{Learning and inference algorithm}
\label{alg:learn}
\begin{algorithmic}
\REQUIRE   \quad \\ (1) training examples $\{x_i\}_{i=1}^n$\\(2) cluster number $K$\\(3) cluster prior probability $\pi=\{\pi_i\}_{i=1}^K$\\(4) number of Langevin steps $l$ and learning iterations $T$
\ENSURE \quad \\ (1) learned parameters $\theta$\\(2) inferred continuous latent variable  $\{z_i\}_{i=1}^n$\\(3) inferred discrete latent variable $\{y_i\}_{i=1}^n$\\~\\

\STATE 1:  Let $t \leftarrow 0$, initialize $\theta$.
\STATE 2:  Initialize $\{z_i\}_{i=1}^n\sim \N(0, I_d)$
\STATE 3:  Initialize $\{y_i\}_{i=1}^n \sim \Cat(\pi)$
\REPEAT
\STATE 4: \textbf{Inference on $z$:} For each observed $x_i$, starting from the current $z_i$ and $y_i$, run Langevin dynamics $l$ steps to update $z_i$ as in Eqn.~\ref{eq:langevin}  \\
\STATE 5: \textbf{Inference on $y$:} For each observed $x_i$, based on the current $z_i$, sample, or obtain a Maximum a Posteriori (MAP), of $y$ using estimated probability as in Eqn.~\ref{eq:posy}. 
\STATE 5: \textbf{Learning $\theta$:} Update $\theta_{t+1} \leftarrow \theta_t+\eta_t L^{\prime}(\theta_t)$, with learning rate $\eta_t$, where $L^{\prime}(\theta_t)$ is computed according to Eqn.~\ref{eq:likelihood_learn}.\\
\STATE 5: Let $t \leftarrow t+1$
\UNTIL{$t=T$}
\end{algorithmic}
\end{algorithm}
\section{Experiments}
\par
In this section, we demonstrate the effectiveness of the proposed model through the experimental results. Firstly, in order to show that the superior unsupervised clustering performance of the proposed model, we provide a quantitative comparison of the unsupervised clustering accuracy of our method with other state-of-the-art methods on three benchmark datasets (i.e. MNIST \cite{lecun1998gradient}, SVHN \cite{netzer2011reading}, STL-10 \cite{coates2011analysis}). Furthermore, to demonstrate that the proposed model can be adapted for inferring 2D label map, we perform unsupervised clustering for per-pixel labels on three datasets (i.e., Facades~\cite{tylevcek2013spatial}, COCO-Stuff~\cite{caesar2018coco} and Potsdam~\cite{isprs4}) and compared it with the CVAE \cite{kingma2014semi} and other state-of-the-art methods. Meanwhile, in order to demonstrate that our proposed model has the ability to learn disentangled latent representations and generate realistic images, we perform image generation experiments on three benchmark datasets. Finally, we also explore the effect of varying $K$'s value on clustering performance. 
\subsection{ Datasets}
\begin{figure}[htbp]
\centering
\subfigure[MNIST]{
\includegraphics[width=0.30\linewidth]{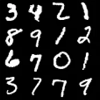}
}
\subfigure[SVHN]{
\includegraphics[width=0.30\linewidth]{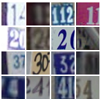}
}
\subfigure[STL-10]{
\includegraphics[width=0.3\linewidth]{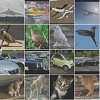}
}
\subfigure[Facades]{
\includegraphics[width=0.3\linewidth]{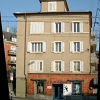}
}
\subfigure[COCO-Stuff]{
\includegraphics[width=0.3\linewidth]{Figure/coco_input.jpg}
}
\subfigure[Potsdam]{
\includegraphics[width=0.3\linewidth]{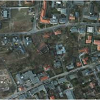}
}
\caption{Some examples of the six datasets.}
\label{fig:dataset}
\end{figure}

To evaluate our method, we use six public datasets: MNIST, SVHN, STL-10, Facades, COCO-Stuff and Potsdam datasets. Figure \ref{fig:dataset} shows an example of these datasets.
\\
\textbf{MNIST:} This is a standard handwritten digits dataset. It consists of 60,000 training samples and 10,000 testing samples. Each image in this dataset consists of $28 \times 28$ pixels, each of which is represented by a gray value. We reshape each image to a 784-dimensional row vector. \\
\textbf{SVHN:} This dataset is obtained from the house number in the Google Street View image. All images in the dataset are $32 \times 32$ color house number images, including 73257 digits for training, 26032 digits for testing sets, and extra 531131 training digits, with approximately 600,000 cropped images. We use testing data to evaluate our unsupervised clustering and rest of the data is used for model training. 

\noindent \textbf{STL-10:} This is an image dataset containing 10 classes of objects, 1,300 per class, 500 training images and 800 testing images. All images in the dataset are $96 \times 96$ color images. We use training images for our model learning and 800 testing images for unsupervised clustering accuracy evaluation.\\
\textbf{Facades:} Facades dataset~\cite{tylevcek2013spatial} is assembled at the Center for Machine Perception, including 606 rectified images of facades from various sources. It is divided into training sets, testing sets and validation sets. The facades are from cities around the world and different architectural styles. We mainly consider four labels including wall, doors, windows and decorations which contains roof, cornice and sill. We need to emphasize that the Facades dataset is commonly used for image-to-image translation \cite{isola2017image} where the image is synthesized given the label map, and in this paper we aim to obtain the label map given the image. \\
\textbf{COCO-Stuff:} COCO-Stuff~\cite{caesar2018coco} is a challenging and diverse segmentation dataset containing ``stuff" classes ranging from buildings to bodies of water. Following the procedure in~\cite{ji2019invariant}, we use the 15 coarse labels
and 52k images variant taking only images with at least 75\% stuff pixels. COCO-Stuff-3 is a
subset of COCO-Stuff with only sky, ground and plants labelled. All input images are shrunk, cropped to $128\times 128$ pixels and Sobel pre-processed as in~\cite{ji2019invariant}.\\
\textbf{Potsdam:} Potsdam~\cite{isprs4} contains 8550 RGBIR $200\times 200$ px satellite images, of which
3150 are unlabelled. As in~\cite{ji2019invariant}, we test the 6-label variant (roads
and cars, vegetation and trees, buildings and clutter) as well as a 3-label variant (Potsdam-3). The construction of Potsdam-3 and the training/testing set preparation also follows~\cite{ji2019invariant}.\\

Note that images from Facades, COCO-Stuff and Potsdam have been manually annotated, however, the annotations are not used in our model training and are only used for ground-truth evaluation.
\begin{figure*}
\subfigure[MNIST]{
\label{fig:a}
\begin{minipage}[b]{0.321\linewidth}
\includegraphics[width=1\textwidth]{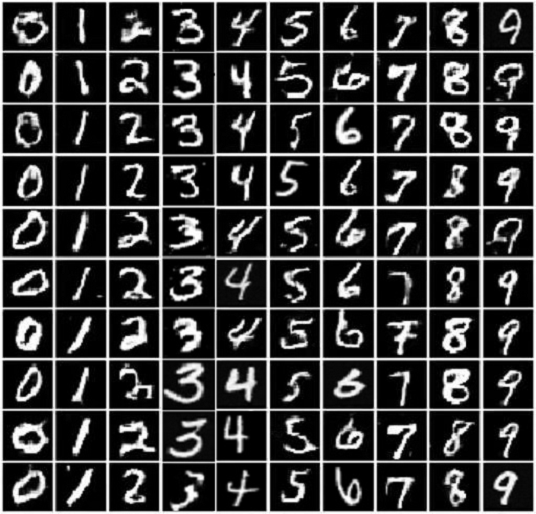}
\end{minipage}
}
\subfigure[SVHN]{
\label{fig:b}
\begin{minipage}[b]{0.321\linewidth}
\includegraphics[width=1\textwidth]{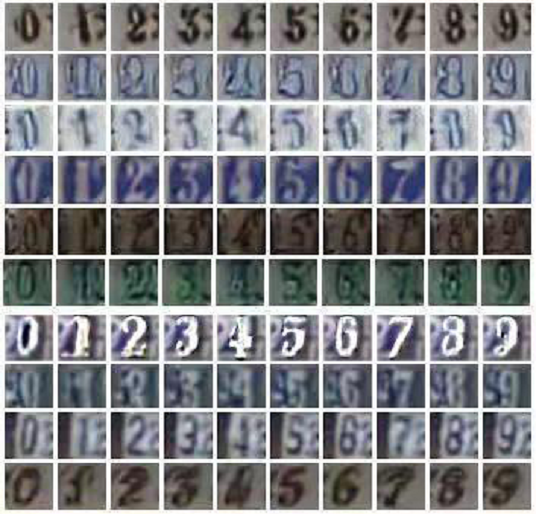}
\end{minipage}
}
\subfigure[STL-10]{
\label{fig:c}
\begin{minipage}[b]{0.321\linewidth}
\includegraphics[width=1\textwidth]{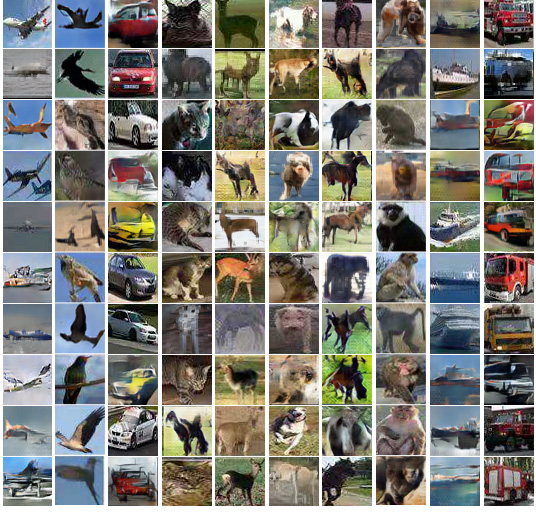}
\end{minipage}
}
\caption{Generated samples by our proposed method. Each row shares the same $z$ and each column shares the same $y$.
(a) Generate samples on the MNIST dataset. (b) Generate samples on the SVHN dataset.  (c) Generate samples on the STL-10 dataset. 
}
\label{fig:digitvisual}
\end{figure*}

\par
\subsection{ Evaluation Metric}
Similar to the work of DEC~\cite{xie2016unsupervised}, we use the unsupervised clustering accuracy (ACC) to evaluate the performance of the proposed method. The formula is defined as follows:
\begin{equation}
\begin{aligned}
ACC = \max \limits_{m \in M} \  \frac {1} {N} \sum \limits_{i=1}^N \1 \{l_{i}=m(c_i)\} ,
\end{aligned}
\end{equation}
where $N$ is the total number of all samples, $l_{i}$ is the ground-truth label and $c_{i}$ is the clustering assignment obtained by various models. $m\in M$ indicates all possible one-to-one mapping set between cluster assignment and labels. KuhnMunkres algorithm~\cite{munkres1957algorithms} is used to find the best mapping.
The range of ACC is between 0 and 1. If the value of ACC is larger, it indicates that the unsupervised classification performance is better.

\par
\subsection{ Implementation Details}
Our implementation is based on Tensorflow \cite{abadi2016tensorflow} framework. The experiments are all carried out on a workstation with NVIDIA GeForce RTX 2080Ti and 1 TB RAM.
\par
During the training process, the parameters of our algorithm are set as follows: we set the standard deviation $\sigma$ of the noise vector $\varepsilon$ to 0.3. In each learning iteration, we set the number of steps $l$ of Langevin dynamic sampling to 100. We performed $T=1000$ learning iterations with learning rate 0.0002 and momentum 0.5. 
\par
The proposed cluster generation model mainly adopts the structure of the deconvolutional-based generator, which is composed of multiple convolutional layers and deconvolution layers. The complete convolutional layer is composed of convolution, ReLU layer and downsampled operation. The deconvolution layer consists of linear superposition, ReLu layer, and upsampling operation. To make the training process more stable, we also use batch normalization~\cite{ioffe2015batch}. The detailed structural information of our proposed clustered generator model will be given later and our experimental code will be released.
\par
We use various convolutional structures to generate the realistic images through our proposed new learning algorithm. Particularly, we mainly introduce the structure of the network for image generation on the MNIST dataset. The network structure for image generation on the other datasets (SVHN, STL-10) is similar to the network structure on the MNIST dataset. Below we describe in detail the structure of the network model for performing image generation on the MNIST dataset as follows.

\par
The proposed network structure consists of 5 layers of convolution and 5 layers of deconvolution layer. In the convolution stage, the convolution kernel size of each layer is $4 \times 4$, the stride from the layer 1 to layer 5 is set to 1, 2, 2, 2, 2, respectively. In the deconvolution stage, the convolution kernel size of each of the deconvolution layer is $4 \times 4$ with stride 2 from layer 6 to layer 9, and the stride on the layer 10 is set to be 1. We utilize the one-hot form of the discrete latent variable with dimension 10, and set dimension for continuous latent variables to be 100. 

\begin{table}[!htbp]
\begin{center}
\begin{tabular}{c|c|c|c|c} \hline
Method    & K     & MNIST     & SVHN     & STL-10  \\ \hline
K-means   &10     & 53.49\%   & 28.40\%     &-- \\
AAE~\cite{makhzani2015adversarial}       &16     & 83.48\%   & 80.01\%     &-- \\
DEC~\cite{xie2016unsupervised}       &10     & 84.30\%    & 80.62\%    & 11.90\%  \\
VaDE~\cite{jiang2016variational}      & 10    & 94.46\%    & 84.45\%   & --\\
HashGAN~\cite{Cao_2018_CVPR}   & 10    & 96.50\%    & 39.40\%    &-- \\
CVAE~\cite{kingma2014semi}      & 10    & 82.26\%    & 62.37\%    & 58.25\% \\
IIC~\cite{ji2019invariant}       & 10    & \textbf{99.2\%}     &--          & 59.6\%\\
\hline
Our method & 10   & 98.35\%  & \textbf{85.15\%}  & \textbf{75.30\%} \\ \hline
\end{tabular}
\end{center}
\caption{ Comparison of unsupervised clustering accuracy (ACC) for various methods on different datasets. }
\label{table1}

\end{table}
\begin{figure}
\centering
\includegraphics[width=0.88\linewidth]{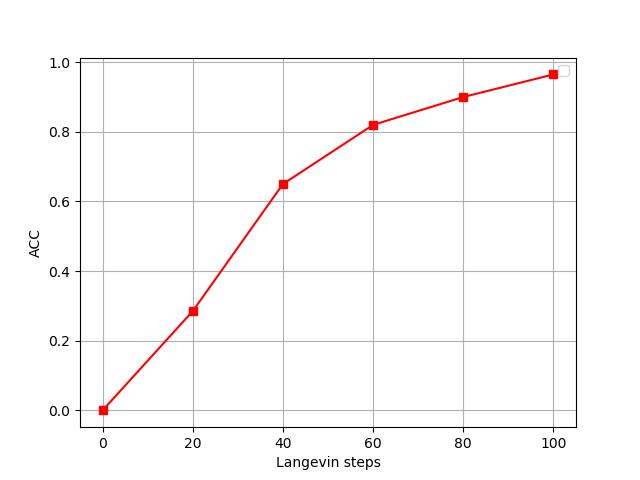}
\caption{The impact of Langevin steps for unsupervised clustering in terms of ACC. More Langevin steps for inference indicate more accurate clustering. }
\label{fig:diff_langevin}
\end{figure}
\par
\subsection{Unsupervised Clustering}
We now evaluate the model on the task of unsupervised clustering. 
We learn our model on the training sets of the benchmark datasets (MNIST, SVHN and STL-10) and evaluate their clustering performance on the corresponding testing sets. Given the test data, we infer its corresponding cluster label using Eqn.~\ref{eq:posy}. If the inference is accurate, then we would expect a competitive unsupervised clustering accuracy as indicated by ACC. We made a quantitative comparison of various clustering methods, and the comparison results are shown in Table \ref{table1}. Note that the CVAE \cite{kingma2014semi} model is primarily developed for supervised/semi-supervised learning settings and we extend it for unsupervised clustering for a fair comparison. As can be seen from Table \ref{table1}, all deep learning models (AAE \cite{makhzani2015adversarial}, DEC \cite{xie2016unsupervised}, VaDE \cite{jiang2016variational}, HashGAN \cite{Cao_2018_CVPR}, IIC\cite{ji2019invariant} and CVAE \cite{kingma2014semi}) perform better than the traditional machine learning methods (K-means\cite{hartigan1979algorithm}). Moreover, we can achieve competitive unsupervised clustering accuracy compared with the state-of-the-art methods. Specifically, on MNIST, SVHN and STL-10 dataset, our method achieves clustering accuracy of 98.35\%, 85.15\% and  75.30\%, which are over the CVAE method by 16.09\%, 12.78\% and 17.05\%, respectively. Performance improvement is more obvious on the STL-10 dataset. 
\par
The competitive or superior clustering accuracy obtained indicates that the inference process of our model is more accurate than the existing variational-based models~\cite{jiang2016variational,makhzani2015adversarial,kingma2014semi}. We argue this is due to the fact that those variational models need carefully designed approximated recognition models $q_\phi(y|x)$ for efficient inference. On the other hand, our model can perform exact inference based on posterior distribution in a unified probabilistic framework which leads to better inference and clustering accuracy. It is worth noting that more steps of Langevin dynamics with Eqn.~\ref{eq:langevin} will render more accurate inference on continuous $z$ which will further improve the accuracy of the unsupervised clustering as can be seen from Figure \ref{fig:diff_langevin}.

\begin{figure*}
\centering

\includegraphics[width=0.21\linewidth]{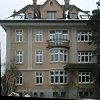}
\hspace{1.2mm}
\includegraphics[width=0.21\linewidth]{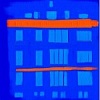}
\hspace{1.2mm}
\includegraphics[width=0.21\linewidth]{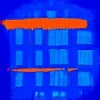}
\hspace{1.2mm}
\includegraphics[width=0.21\linewidth]{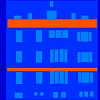}\\

\vspace{0.9mm}
\includegraphics[width=0.21\linewidth]{Figure/coco_new/input.png}
\hspace{1.2mm}
\includegraphics[width=0.21\linewidth]{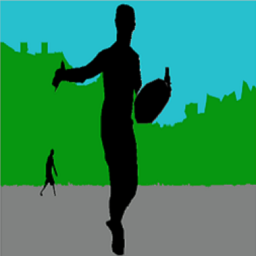}
\hspace{1.2mm}
\includegraphics[width=0.21\linewidth]{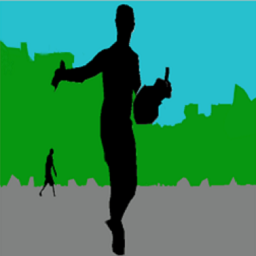}
\hspace{1.2mm}
\includegraphics[width=0.21\linewidth]{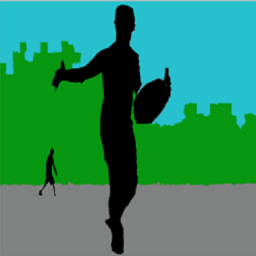}\\

\vspace{0.9mm}
\includegraphics[width=0.21\linewidth]{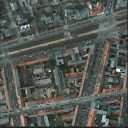}
\hspace{1.2mm}
\includegraphics[width=0.21\linewidth]{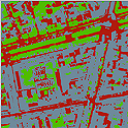}
\hspace{1.2mm}
\includegraphics[width=0.21\linewidth]{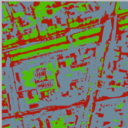}
\hspace{1.2mm}
\includegraphics[width=0.21\linewidth]{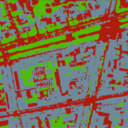}\\
\caption{Qualitative comparison of our clustering results with the CVAE method on three datasets: Facade (top), COCO-Stuff-3 (middle) and Potsdam-3 (bottom). The first column is the testing image. The second column is the clustering result of our method. The third column is the clustering result of the CVAE method, and the ground-truth (GT) label map is shown in the last column. COCO-Stuff-3 considers labels: sky, vegetation and ground. The Potsdam-3 considers: vegetation, roads and buildings.}
\label{fig:per_pixel}
\end{figure*}

\par
\subsection{Per-pixel Unsupervised Clustering }
In this section, we evaluate the ability of the model to accurately infer 2D discrete latent map by performing unsupervised clustering tasks on three datasets: Facades, COCO-stuff and Potsdam. To the best of our knowledge, there is currently among the only few methods~\cite{ji2019invariant} that attempt to perform per-pixel unsupervised clustering of an image. The main challenge is that the per-pixel clustering should conform to the underlying pixel-wise relations (e.g., consistency for neighbouring regions) which require accurate inference. Our proposed model can obtain reasonably well per-pixel unsupervised clustering result. 
\begin{table}[!htbp]
\begin{center}
\resizebox{1.0\columnwidth}{!}{
\begin{tabular}{c|c|c|c|c} \hline
Method    & COCO-Stuff-3     & COCO-Stuff    & Potsdam-3     & Potsdam  \\ \hline
K-means   &52.2\%     & 14.1\%   & 45.7\%     &35.3\% \\
SIFT~\cite{lowe2004distinctive} &38.1\% &20.2\% & 38.2\% &28.5\%\\
DeepCluster~\cite{caron2018deep} &41.6\% & 19.9\% &41.7\% & 29.2\%\\ 
Co-Occurrence~\cite{isola2015learning} &54.0\% &24.3\% &63.9\% &44.9\%\\
IIC~\cite{ji2019invariant}       &72.3\% & 27.7\% & 65.1\% & 45.4\% \\
CVAE~\cite{kingma2014semi} & 62.4\%&24.5\% &61.9\% & 39.8\%\\
\hline
Our method & \textbf{73.3\%}   & \textbf{28.1\%}  & \textbf{66.3\%}  & \textbf{46.2\%} \\ \hline
\end{tabular}
}
\end{center}
\caption{ Comparison of unsupervised clustering accuracy (ACC) for various methods on different datasets. The accuracy numbers except CVAE and our model are from~\cite{ji2019invariant}. }
\label{table2}
\end{table}
\par
Unlike tradition clustering methods, we cluster each pixel on the label map. The traditional clustering method, as we show in the previous experiment, considers one-dimensional vector space which forms a one-hot representation. It should be noted that we are now performing clustering in a two-dimensional pixel space. Specifically, we consider one-hot representation for every pixel based on which we perform the  inference using Eqn.~\ref{eq:infery}. In order to make a fair comparison with the CVAE model, all other settings (e.g. the number of labels in datasets and the number of iterations of the unsupervised clustering algorithm) are kept untouched except for the clustering method.  


For a qualitative comparison, we present the cluster assignment obtained by our method and CVAE method in the form of label maps and compare them with the ground truth labels. The visualization of the per-pixel unsupervised clustering results is shown in Figure \ref{fig:per_pixel}. We also quantitatively compare with the CVAE and other related baseline models in terms clustering accuracy (ACC) on COCO-Stuff and Potsdam datasets. The preparation of the datasets are followed by the routine in~\cite{ji2019invariant} and the results are shown in Table~\ref{table2}. Note that the baseline models (SIFT~\cite{lowe2004distinctive}, DeepCluster~\cite{caron2018deep},Co-Occurrence~\cite{isola2015learning}  ) do not directly learn a clustering function and requires further application of k-means to be used for image clustering. The most recent IIC~\cite{ji2019invariant} model  can directly learn 2D clustering map, however, it only learns the feature embedding and is unable to represent the observed data distribution, therefore does not have generation ability as we do in Sec.\ref{sec:generation}.

As shown in Figure \ref{fig:per_pixel}, compared to CVAE method, our approach can better preserve the internal structure of the building and objects, and can also clearly display the details. This can be further verified by Table~\ref{table2} where our model achieve the competitive or better clustering accuracy. 



\subsection{Image Generation}
\label{sec:generation}
Our model can not only obtain the powerful data embedding to ensure the accurate unsupervised clustering, it can also learn the disentangled latent representations to generate realistic samples. 
To demonstrate the effectiveness of our proposed, we perform experiments on the MNIST, SVHN and STL-10 datasets. We set $K=10$ on three datasets to train our proposed model and show that the learning and inference of the latent variables could obtain disentangled latent representations of the data. To show this, we obtain the generated samples through learned clustered generator model by varying the two sets of latent variables in the following way:
\par
(1) Firstly, we change the continuous latent variable $z$ within a certain range by fixing the discrete class $y$;
\par
(2) Secondly, we fix the continuous latent variable $z$ and enumerate all possible values of discrete class label $y$. 
\par
Figure \ref{fig:digitvisual} shows the generation result of our model on the three datasets MNIST, SVHN and STL10. As can be seen from Figure \ref{fig:digitvisual}, 
the image generated by our model is both realistic and diverse. Meanwhile, it can be clearly seen that if the cluster label is fixed, the generated samples have different styles and variations while maintaining their identity, indicating that continuous latent variable $z$ effectively captures the variations within each cluster.  On the other hand, the change of discrete latent variable $y$ could change the identity of the sample, indicating that it can be effective for cluster label modeling. Therefore, the learned discrete latent variable $y$ and continuous $z$ form the disentangled latent representation. 

\begin{figure}
\centering
\subfigure[k=6 ]{
\label{fig:a our method}
\begin{minipage}[b]{0.46\linewidth}
\includegraphics[width=1\textwidth]{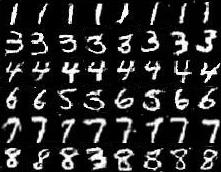}
\end{minipage}
}
\subfigure[k=12]{
\label{fig:b}
\begin{minipage}[b]{0.46\linewidth}
\includegraphics[width=1\textwidth]{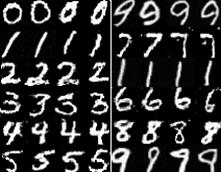}
\end{minipage}
}
\caption{ Visual comparison of the clustering results by setting different number of clusters (i.e. 6 and 12) on the MNIST dataset.}
\label{fig:clusters_k}
\end{figure}

\par
\subsection{The Impact of the Number of Clusters }
The number of clusters $K$ is given as priori in our model, and $K$ is set to be the number of classes for each dataset. To further investigate how different $K$ could affect our model, we conduct experiments on the MNIST dataset for different $K$. We randomly set different $K$ values on the MNIST dataset, such as 6 and 12. The experimental results of clustering are shown in Figure \ref{fig:clusters_k}. It can be seen from Figure \ref{fig:clusters_k} that if the number of clusters $K$ is smaller than the actual number of classes, digits with similar appearances are grouped together, such as digits 3, 6, and 5. If the number of clusters $K$ is larger than the actual the number of classes, some digits are divided into subclasses based on visually appearance identifiable attributes, such as digits italics and roundness. As can be seen from the Figure \ref{fig:clusters_k} \subref{fig:b}, the upright and oblique 1 are divided into two clusters, and the 9 with two handwritten styles are also divided into two clusters.

\section{Conclusion}

In this paper, we propose the clustered generator model for the task of unsupervised clustering. The clustered generator model contains both the discrete latent variables which capture the cluster labels and the continuous latent variables which capture the variations within the clusters. We then develop the novel learning and inference algorithm for clustered generator in a unified probabilistic framework. Specifically, we iteratively infer the continuous and discrete latent variables in a Gibbs manner, then use the inferred variables to learn the clustered generator model. The learning can naturally incorporate the unsupervised clustering as an inference step without the need for extra assisting models for approximation. The latent variables learned can be served as both observed data embedding as well as latent representations for data distribution. The extensive experiments show both quantitatively and qualitatively the effectiveness of our proposed model. 

The model can be adapted for semi-supervised learning given only a small portion of the label. The model can also be generalized to a dynamic one by including the transition model for latent variables. Besides, the number of clusters $K$ is pre-specified in the current work and can be learned directly from data. We leave these as our future directions. 

\section*{Acknowledgment}
The work is partially supported by DARPA XAI project N66001-17-2-4029. 

{\small
\bibliographystyle{ieee_fullname}
\bibliography{mybibfile}
}

\end{document}